\documentclass[10pt,twocolumn,letterpaper]{article}

\usepackage[pagenumbers]{cvpr} % To force page numbers, e.g. for an arXiv version

\usepackage[dvipsnames]{xcolor}

\usepackage{graphicx}
\usepackage{amsmath}
\usepackage{amssymb}
\usepackage{booktabs}

\graphicspath{{images}}

\definecolor{cvprblue}{rgb}{0.21,0.49,0.74}
\usepackage[pagebackref,breaklinks,colorlinks,citecolor=cvprblue]{hyperref}

\def\paperID{*****} % *** Enter the Paper ID here
\def\confName{CVPR}
\def\confYear{2024}

\title{Multi-Scale Semantic Segmentation with Modified MBConv Blocks}

\author{Xi Chen \qquad Yang Cai \qquad Yuan Wu \qquad Bo Xiong \qquad Taesung Park \\
Department of Computer Science, Princeton University\\
{\tt\small \{xichen, yangcai\}@cs.princeton.edu}
}

\begin{document}
\definecolor{Green}{RGB}{0, 170, 0}
\definecolor{Red}{RGB}{200, 0, 0}

\maketitle

\begin{abstract}
Recently, MBConv blocks—initially designed for efficiency in resource-limited settings and later adapted for cutting-edge image classification performances—have demonstrated significant potential in image classification tasks. Despite their success, their application in semantic segmentation has remained relatively unexplored. This paper introduces a novel adaptation of MBConv blocks specifically tailored for semantic segmentation. Our modification stems from the insight that semantic segmentation requires the extraction of more detailed spatial information than image classification. We argue that to effectively perform multi-scale semantic segmentation, each branch of a U-Net architecture, regardless of its resolution, should possess equivalent segmentation capabilities. By implementing these changes, our approach achieves impressive mean Intersection over Union (IoU) scores of 84.5\% and 84.0\% on the Cityscapes test and validation datasets, respectively, demonstrating the efficacy of our proposed modifications in enhancing semantic segmentation performance.
\end{abstract}

\section{Introduction}
Deep convolutional neural networks have set new benchmarks across a wide array of computer vision applications, including image classification, object detection, semantic segmentation, and human pose estimation. Semantic segmentation, in particular, involves the precise categorization of each pixel within an image into specific class labels, offering a comprehensive analysis of the scene that encompasses the prediction of the label, location, and shape of every element. This field has garnered widespread attention due to its potential to revolutionize areas such as autonomous driving and robotic sensing, among others, by providing detailed and actionable insights into the surrounding environment.

\subsection{MBConv Blocks}
Recently, MBConv blocks \cite{sandler2018mobilenetv2}, characterized by inverted residual structures with linear bottlenecks, have achieved leading-edge accuracy in image classification tasks. These blocks are ingeniously crafted to optimize performance, even in scenarios with limited computational resources. The architecture of MBConv blocks incorporates three key components to realize this high level of efficiency and accuracy: Depthwise Separable Convolutions, Linear Bottlenecks, and Inverted Residuals. Each component plays a pivotal role in enhancing the network's ability to process and learn from image data effectively, making MBConv blocks a cornerstone for resource-efficient, high-accuracy image classification models.
\par
Depthwise Separable Convolutions serve as foundational elements in numerous high-efficiency network architectures, thanks to their streamlined computational model \cite{sandler2018mobilenetv2,howard2017mobilenets,chollet2017xception}. These blocks are composed of two integral operations: a depthwise convolution, which employs a distinct convolutional filter for each input channel, and a pointwise convolution, a $1 \times 1$ convolution that synthesizes new features through linear combinations of the input channels. This dual-step process significantly enhances computational efficiency, accelerating the network's performance without compromising the quality of the output.
\par
Linear Bottlenecks embody a dual-pronged concept: firstly, that feature maps can be effectively compressed into low-dimensional subspaces without significant loss of information, and secondly, that the application of nonlinear activations may lead to information degradation. In the bottleneck architecture described in earlier works \cite{he2016deep,khoshsirat2023improving,khoshsirat2023sentence}, the process begins by mapping the input to a reduced dimensionality, where it is then processed, maintaining the feature representation within this compact space. Conversely, the inverted bottleneck approach flips this paradigm by retaining features in the low-dimensional space while conducting the bulk of processing in an expanded, higher-dimensional context. This innovative strategy optimizes information flow and processing efficiency within neural network architectures.
\par
In this study, we propose a hypothesis that underscores the critical need for semantic segmentation networks to extract precise spatial context for each pixel, a requirement that starkly contrasts with the demands of image classification networks. While classification networks focus on extracting features sufficient for categorizing images, thereby negating the necessity for maintaining pixel-specific spatial accuracy, semantic segmentation tasks demand a more nuanced approach. Motivated by this distinction, we have tailored modifications to the MBConv blocks, enabling them to capture an enhanced spatial context, thus significantly improving their efficacy in semantic segmentation applications.
\subsection{Multi-Scale Segmentation}
Despite all images in a dataset sharing the same resolution, the scale of objects within these images varies significantly. This variation necessitates performing image segmentation at multiple scales, as objects from any class can be present at any scale. Certain methodologies, as referenced in studies \cite{ronneberger2015u,wang2020deep,chen2018encoder}, deliberately amalgamate features from various scales to achieve the final segmentation result. Conversely, other approaches \cite{zhao2017pyramid,badrinarayanan2017segnet,long2015fully} rely solely on features from the final scale—which typically has the lowest resolution—for constructing the segmentation map, using higher resolution features merely as transitional steps in the process. This delineation highlights the diverse strategies employed to address the challenges posed by scale variation in image segmentation tasks.
\par
In this paper, we show that it is crucial to explicitly use the higher resolution features, and the higher resolution branches should have the same segmentation and classification power as the lower resolution branches.

\begin{figure*}[t]
\centering
\includegraphics[width=1.0\linewidth]{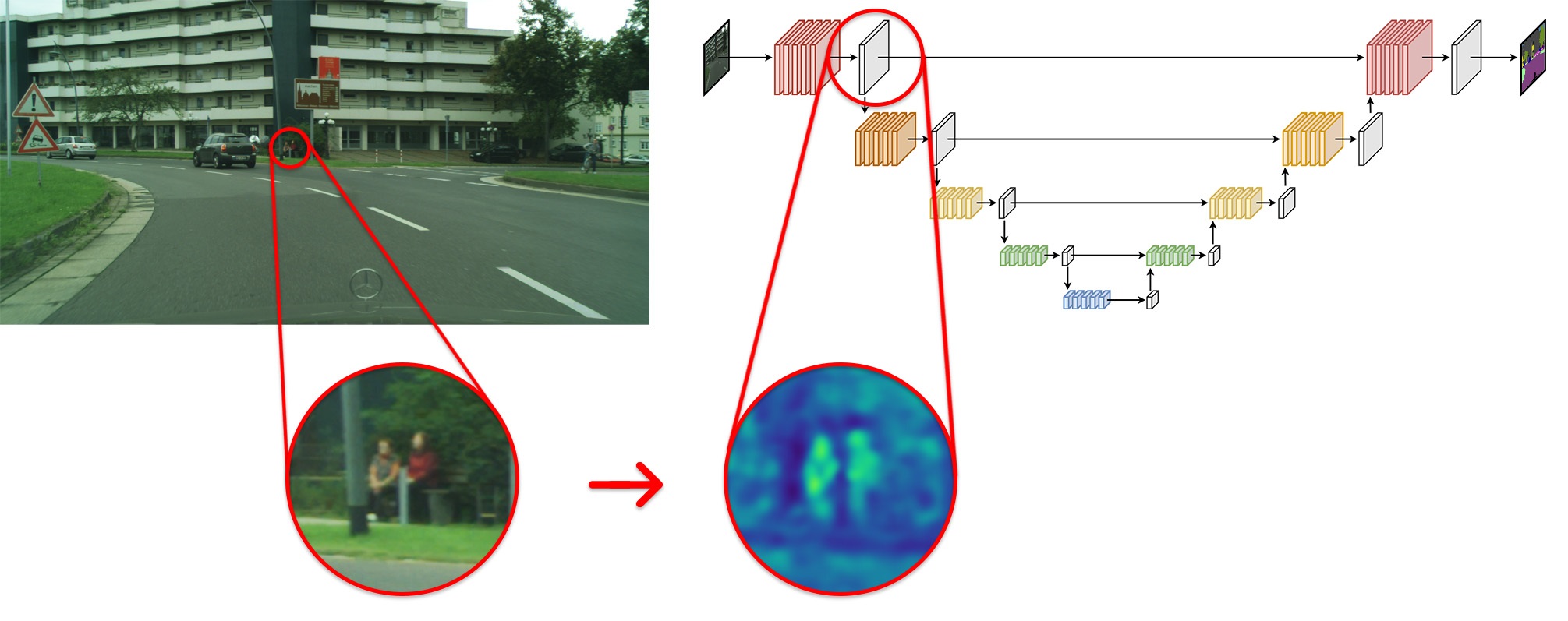}
\caption{
The higher resolution feature maps show that these branches are able to segment the smaller objects in the images (the context can affect the final class). This observation shows that the higher resolution branches need to have the same learning power as the lower resolution ones, since they need to classify and segment similar number of classes and objects.
}
\label{fig:figure1}
\end{figure*}

\section{Related Work}
Current advancements in semantic segmentation prominently feature convolutional neural networks (CNNs) with varied architectures tailored for specific computer vision tasks, including object detection \cite{lin2017feature,khoshsirat2023empowering}, human pose estimation \cite{newell2016stacked,khoshsiratembedding}, image-based localization \cite{melekhov2017image,khoshsirat2023transformer,maserat201743}, and notably, semantic segmentation \cite{long2015fully,badrinarayanan2017segnet,noh2015learning}. Among these, encoder-decoder or hourglass architectures are prevalent, designed with an encoder that progressively compresses feature maps to distill high-level semantic content, and a decoder that incrementally restores low-level details. However, the inherent reduction in image detail during encoding means these networks typically cannot achieve optimal performance without incorporating skip connections, as exemplified by the U-Net \cite{ronneberger2015u}, which leverages feature maps from the encoder to recapture fine image particulars.
\par
Further diversifying the landscape, spatial pyramid pooling models, such as PSPNet \cite{zhao2017pyramid} and DeepLab \cite{chen2017rethinking}, integrate spatial pyramid pooling \cite{lazebnik2006beyond,grauman2005pyramid} across varying grid scales or utilize multiple atrous convolutions \cite{chen2017deeplab} at different rates to enrich feature representation. DeepLabv3+ \cite{chen2018encoder} enhances this approach by adding a skip connection to preserve some low-level image nuances. Meanwhile, high-resolution representation networks \cite{wang2020deep,huang2017multi,fourure2017residual,zhou2015interlinked} aim to maintain a high-resolution state throughout the processing chain, extracting high-level semantics without sacrificing low-level details through parallel streams of low-resolution convolutions. However, to manage the substantial memory demand, these models initially reduce the input image's resolution before proceeding with the primary computational processes.
\par
Several strategies \cite{chen2017deeplab,chandra2016fast,khoshsirat2022semantic} employ post-processing techniques like conditional random fields to refine the output of neural networks, enhancing segmentation precision particularly around object edges. Although effective, these methods introduce additional computational load during both training and testing phases. In contrast, pyramid pooling approaches generally capture context within square regions, utilizing pooling and dilation in a symmetric manner. Relational context methods, on the other hand, diverge from this geometric constraint by focusing on the inter-pixel relationships, thus enabling context analysis beyond mere square regions. This adaptability allows for more tailored context understanding in complex semantic landscapes, such as dispersed areas or elongated structures.
\par
Innovative networks like OCRNet \cite{yuan2020object}, DANet \cite{xue2019danet}, and CFNet \cite{zhang2019co} push the boundaries further by enhancing pixel representation through the aggregation of contextual pixel information, where the context is defined by the entirety of pixels within an image. These methods leverage the concept of self-attention \cite{wang2018non,vaswani2017attention,hosseini2022application}, considering the relational dynamics between pixels, and utilize a weighted aggregation approach where the weights are determined by pixel similarities. Such methodologies serve as valuable enhancements to conventional segmentation frameworks, offering a nuanced layer of contextual analysis that can significantly improve segmentation outcomes.
\par
The concept of multi-scale image segmentation, recognizing the presence of objects at varying scales within images, has been a cornerstone in segmentation research for many years \cite{tabb1997multiscale,vincken1997probabilistic,choi2001multiscale}. These methods underscore the necessity of performing segmentation at multiple scales to accurately identify and delineate objects of different sizes. However, a significant challenge arises from the high computational demand associated with processing images at elevated resolutions. This limitation has led contemporary state-of-the-art techniques to adopt strategies that minimize computational expenditure at higher resolution levels, often at the cost of detailed accuracy.
\par
Recent advancements in this field are predominantly anchored in the capabilities of convolutional neural networks. Notably, the work presented in \cite{zhao2019multi} adopts a deep supervision mechanism, aiming to mitigate the computational challenges while enhancing the effectiveness of multi-scale segmentation. This approach represents a strategic effort to balance resource utilization with the need for precision across different scales, highlighting the ongoing evolution and optimization of segmentation methods in response to the inherent challenges of multi-scale image analysis.
\par
In the exploration of enhanced segmentation models, \cite{zhou2018unet++} introduces a variation of the U-Net architecture augmented with additional skip connections and deep supervision to refine feature integration across different network depths. Meanwhile, \cite{schmitz2021multi} employs multi-scale fusion techniques within a modified U-Net framework to effectively encapsulate global contextual information, demonstrating the potential of architectural modifications in improving segmentation performance. However, it's noted that these innovative approaches are seldom applied to prominent semantic segmentation datasets, highlighting a gap in their widespread validation and adoption.
\par
In parallel, the concept of self-training has shown promise in enhancing classification networks, as illustrated by \cite{yalniz2019billion}. A notable advancement is made in \cite{xie2020self}, where the Noisy Student algorithm is leveraged to set a new benchmark on the ImageNet dataset \cite{russakovsky2015imagenet}. This technique's utility is further evidenced in the Cityscapes dataset, where the inherent coarseness of labels leaves substantial portions of images unlabeled. The study in \cite{tao2020hierarchical} successfully applies Noisy Student Training within a multi-module strategy introduced by \cite{yuan2020object} to surmount the challenges posed by sparse labels, thereby boosting segmentation accuracy.
\par
Drawing from the strengths and weaknesses of these diverse methodologies, our current work proposes the development of an independent network design. This novel network processes images at their native, high-resolution state and directly generates high-resolution segmentation maps. By focusing on maintaining the original image quality throughout the processing pipeline, this approach aims to achieve superior segmentation accuracy with enhanced generalization capabilities, addressing a critical need for effective high-resolution image analysis in semantic segmentation tasks.

\begin{figure*}[t]
\centering
\includegraphics[width=1.0\linewidth]{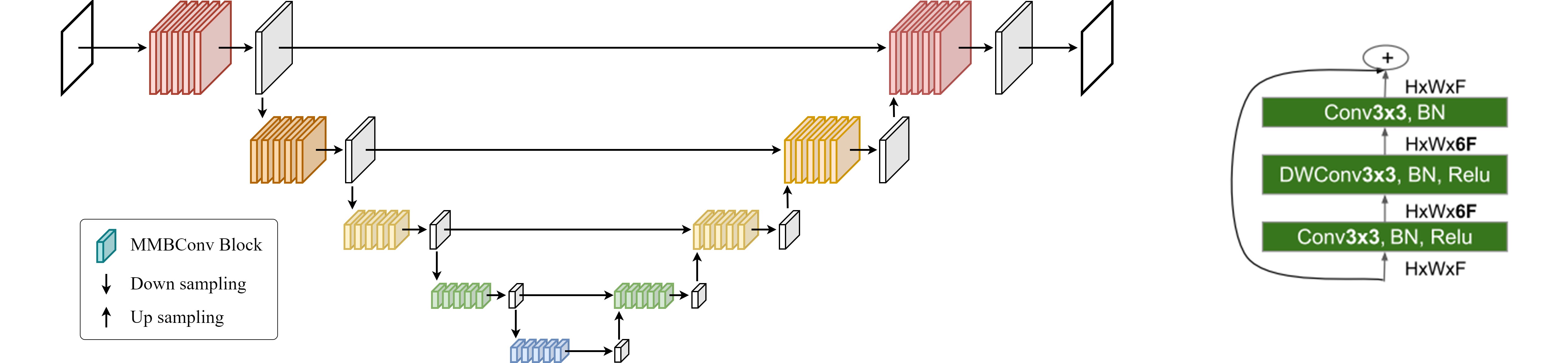}
\caption{
The proposed modifications. \textbf{Left}: Our modified U-Net. All the branches have the same depth and number of channels. The residual blocks are replaced with our modified MBConv blocks. \textbf{Right}: Our modified MBConv block. The $1 \times 1$ convolutions are replaced with $3 \times 3$ convolutions.
}
\label{fig:figure2}
\end{figure*}

\section{Method}
In this study, we propose two significant enhancements to the widely recognized U-Net architecture \cite{ronneberger2015u}. For each of these modifications, we conduct a detailed comparative analysis, juxtaposing our results with those of analogous methodologies reported in the existing literature. This approach enables us to precisely evaluate the efficacy of our improvements within the context of the broader research landscape, highlighting their potential to advance the state-of-the-art in segmentation technology.

\subsection{Multi-Scale Segmentation}
In the realm of semantic segmentation, current network designs typically employ a reduced number of feature maps at higher resolutions, a compromise necessitated by computational constraints. Contrary to this trend, our approach advocates for maintaining a consistent number of feature maps and architectural blocks across all scales. This strategy is predicated on the understanding that the task of segmenting and classifying objects across various scales demands uniform computational capability, reflected here by the consistent allocation of feature maps. As illustrated in Figure \ref{fig:figure1}, the highest resolution branch of a U-Net can independently segment smaller objects within an image, demonstrating that context from other branches does not significantly influence the segmentation outcome.
\par
To refine the U-Net architecture in line with our proposition, we ensure that all branches feature equivalent depth and number of channels. Addressing potential concerns regarding increased memory usage, we subtly augment the channel count in higher resolution branches while decreasing it in lower resolution counterparts. Furthermore, we introduce a stem module to the network, effectively reducing the input resolution by a factor of four. This balanced approach aims to harness the strengths of uniform learning power across scales while mitigating memory overhead, setting the stage for more efficient and effective semantic segmentation.

\begin{table}[t]
\centering
\setlength\tabcolsep{15pt}
\begin{tabular}{l|c}
Method & Mean IoU \\
\hline
HRNetV2 + OCR (w/ ASP) \cite{yuan2020object} & 83.67 \\
DecoupleSegNet \cite{li2020improving} & 83.70 \\
EfficientPS \cite{mohan2021efficientps} & 84.24 \\
HRNet + OCR + SegFix \cite{yuan2020object} & 84.50 \\
Panoptic-DeepLab \cite{cheng2020panoptic} & 84.54 \\
\hline
\textbf{Ours (MMBConv)} & \textbf{84.58} \\
\hline
\end{tabular}
\caption{State-of-the-art results on the Cityscapes \cite{cordts2016cityscapes} test set for different network architectures.}
\label{tab:table1}
\end{table}

\subsection{Modified MBConv Blocks}
The MBConv blocks, characterized by inverted residuals and linear bottleneck structures \cite{sandler2018mobilenetv2}, have become a staple in the realm of classification tasks \cite{tan2019efficientnet,xie2020self,cubuk2020randaugment}. When these blocks are integrated into the U-Net architecture in place of traditional residual blocks, only a marginal improvement in accuracy is observed. This outcome is attributed to the intrinsic design of classification networks, which are primarily focused on extracting as many features as possible without necessarily preserving detailed spatial information for each pixel. Conversely, segmentation networks demand precise spatial context to accurately delineate the shape and class of segmented objects.
\par
The challenge lies in enhancing the network's ability to capture spatial details without exponentially increasing computational demands. Our proposed solution is to substitute all $1 \times 1$ convolutions within these blocks with $3 \times 3$ convolutions. While $1 \times 1$ convolutions are traditionally employed for feature mapping, switching to $3 \times 3$ convolutions enables the network to capture more spatial information in addition to performing the mapping function. This alteration leads to an approximate increase in memory usage by 10\% and processing time by 30\%. Figure \ref{fig:figure2} provides a detailed visual representation of these modified blocks, illustrating how this strategic change facilitates a more nuanced understanding of spatial context, potentially improving segmentation accuracy without the need for significantly deeper network architectures.

\begin{figure*}[t]
\centering
\includegraphics[width=1.0\linewidth]{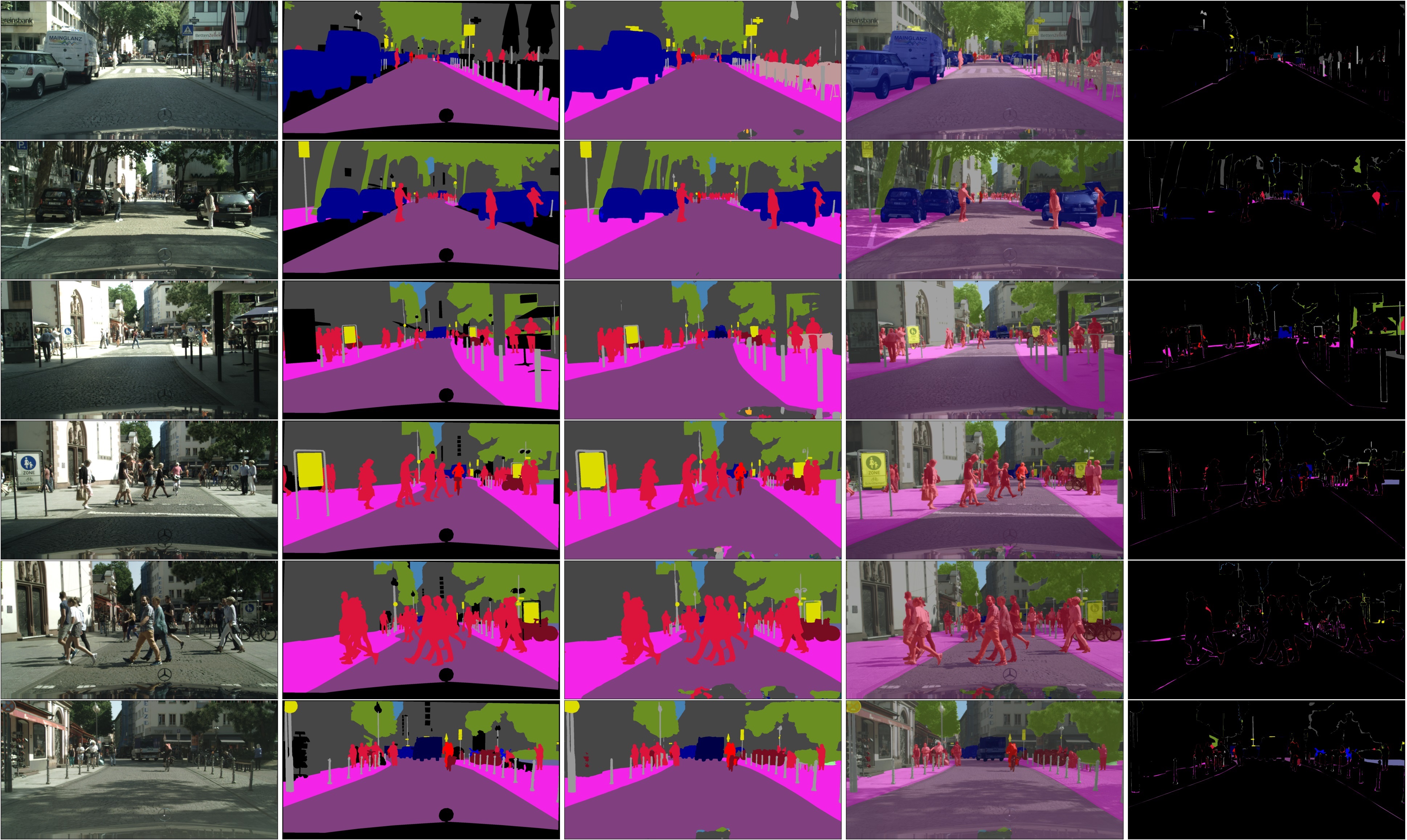}
\caption{
Sample qualitative results from the Cityscapes \cite{cordts2016cityscapes} validation set. From left to right: input image, ground truth, prediction, prediction overlaid on the input image, and the segmentation error.
}
\label{fig:figure3}
\end{figure*}

\begin{table}[t]
\centering
\setlength\tabcolsep{15pt}
\begin{tabular}{l|c}
Method & Mean IoU \\
\hline
EfficientPS \cite{mohan2021efficientps} & 82.1 \\
HRNetV2 + OCR \cite{yuan2020object} & 82.4 \\
Panoptic-DeepLab \cite{cheng2020panoptic} & 83.1 \\
DecoupleSegNet \cite{li2020improving} & 83.5 \\
\hline
\textbf{Ours (MMBConv)} & \textbf{84.0} \\
\hline
\end{tabular}
\caption{State-of-the-art results on the Cityscapes \cite{cordts2016cityscapes} validation set for different network architectures.}
\label{tab:table2}
\end{table}

\section{Experiments}
In our methodology, the encoder segment of our network undergoes initial pretraining on the ImageNet dataset \cite{russakovsky2015imagenet}, leveraging its vast and diverse range of images to capture a broad spectrum of features. Subsequently, we further pretrain the network on the Mapillary Vistas dataset \cite{neuhold2017mapillary}, enriching its capability to interpret complex urban scenes. This preparatory phase is crucial before proceeding to the final training and evaluation stages on the Cityscapes dataset \cite{cordts2016cityscapes}, which focuses on urban street scenes for semantic segmentation tasks.
\par
For optimization, we employ the Lookahead optimizer \cite{zhang2019lookahead} in conjunction with RAdam, optimizing the process with a weight decay set at 0.0001 and a batch size of 16. To effectively manage the learning rate, we adopt a polynomial learning rate policy, setting the poly exponent at 0.9 and the initial learning rate at 0.001. This nuanced approach to learning rate adjustment plays a pivotal role in gradually reducing the learning rate, thereby ensuring more stable convergence over training iterations.
\par
To enhance the network's generalization ability across diverse urban scenes, we implement synchronized batch normalization, utilizing multiple GPUs to normalize the batch statistics. This technique is particularly beneficial for maintaining consistency in the network's performance across different data distributions.
\par
Data augmentation strategies, including random cropping, scaling within a range of [0.5, 2.0], and random horizontal flipping, are applied to introduce variability and robustness in the training process. These augmentation techniques simulate a variety of perspectives and scales, further enhancing the network's adaptability and performance in real-world urban environments.

\subsection{Mapillary Vistas}
The Mapillary Vistas dataset (research edition) \cite{neuhold2017mapillary} represents a comprehensive collection of street-level imagery, meticulously annotated to support a wide range of computer vision tasks. It encompasses approximately 25,000 images, thoughtfully divided into subsets of 18,000 for training, 2,000 for validation, and 5,000 for testing. This dataset is distinguished by its rich diversity, featuring 65 distinct object categories alongside a 'void' class for unclassifiable elements. Moreover, it accommodates a variety of image dimensions, with aspect ratios and resolutions extending up to 22 Megapixels, providing a robust challenge for semantic segmentation algorithms due to the high level of detail and complexity in the scenes.
\par
For our project, we leverage both the training and validation segments of the Mapillary Vistas dataset during the pre-training phase. This approach ensures that our model is exposed to a broad and challenging array of urban scenes and object interactions, facilitating a more comprehensive understanding and interpretation of complex urban environments. The diversity and scale of the Mapillary Vistas dataset make it an invaluable resource for advancing the performance of semantic segmentation models, particularly those tasked with interpreting the nuanced and variable nature of street-level imagery.

\subsection{Cityscapes}
The Cityscapes dataset \cite{cordts2016cityscapes} is a pivotal resource in the field of semantic segmentation, featuring 5,000 high-resolution street images with precise pixel-level annotations. These finely annotated images are systematically allocated into training, validation, and testing sets, consisting of 2,975, 500, and 1,525 images respectively. In addition to these meticulously detailed images, Cityscapes also offers an extensive collection of 20,000 images with coarser annotations, providing a broader base for model training and evaluation.
\par
Cityscapes categorizes urban scene elements into 30 distinct classes, out of which 19 are designated for performance evaluation. This selective focus enables a concentrated assessment on classes that are most relevant to urban street environments. To maximize the accuracy of our model on the test set, we incorporate not just the finely annotated training and validation images but also the coarsely annotated dataset during our training process. This comprehensive training strategy, leveraging the full spectrum of available data, is designed to enhance the model's predictive accuracy and its ability to generalize across a wide array of urban scenes, thereby setting a robust foundation for advanced semantic segmentation tasks.

\subsection{Results}
In our study, we benchmark the performance of our network architecture against a range of existing models, focusing on semantic segmentation accuracy within the Cityscapes dataset \cite{cordts2016cityscapes}. The comparative results are systematically presented in Table \ref{tab:table1} for the test set and Table \ref{tab:table2} for the validation set of Cityscapes, showcasing our approach's superior accuracy across both datasets in comparison to alternative network designs. Figure \ref{fig:figure3} shows sample qualitative results.
\par
A notable strength of our method lies in its simplicity and architectural elegance. Unlike some state-of-the-art solutions that rely on complex, multi-modular designs or intricate segmentation heads, our network architecture is streamlined and straightforward. This design philosophy not only facilitates easier implementation and adaptability across various platforms and tasks but also simplifies the modification process to meet specific requirements.
\par
The comparative ease of understanding and implementing our model stands in contrast to more convoluted approaches, which often pose significant challenges in terms of interpretability and practical application. By achieving high accuracy without the need for excessive complexity, our approach demonstrates that efficiency and effectiveness in semantic segmentation can be attained through thoughtful, minimalist design, making it a valuable addition to the field and a robust foundation for future innovations.

\begin{table}[t]
\centering
\setlength\tabcolsep{15pt}
\begin{tabular}{l|c}
Method & Mean IoU \\
\hline
Baseline U-Net & 76.3 \\
Zhao et al. \cite{zhao2019multi} & 76.5 \\
Schmitz et al. \cite{schmitz2021multi} & 76.6 \\
Unet++ \cite{zhou2018unet++} & 76.8 \\
\hline
\textbf{Our multi-scale approach} & \textbf{77.1} \\
\hline
\end{tabular}
\caption{Comparison of our multi-scale approach with some of the existing methods on the Cityscapes \cite{cordts2016cityscapes} validation set. Multi-scale inference is used.}
\label{tab:table3}
\end{table}

\begin{table}[t]
\centering
\setlength\tabcolsep{15pt}
\begin{tabular}{l|c}
Method & Mean IoU \\
\hline
Baseline U-Net & 76.3 \\
U-Net with MBConv blocks \cite{sandler2018mobilenetv2} & 76.8 \\
\hline
\textbf{U-Net with MMBConv blocks} & \textbf{77.4} \\
\hline
\end{tabular}
\caption{The effect of our modification to the MBConv blocks on the Cityscapes \cite{cordts2016cityscapes} validation set. Multi-scale inference is used.}
\label{tab:table4}
\end{table}

\section{Ablation studies}
In our ablation studies, we establish a baseline using an enhanced version of U-Net, augmented with residual blocks and deeper network branches. This design choice is intentional, aiming to ensure that all network configurations being tested are aligned in terms of computational resource consumption. This allows for a fair and direct comparison of performance impacts resulting from various architectural modifications. To evaluate the effectiveness of these configurations, we utilize the Cityscapes dataset \cite{cordts2016cityscapes}, specifically its validation set. This approach enables us to meticulously assess the impact of each modification on the model's performance, ensuring that any observed improvements in segmentation accuracy are attributable to the architectural changes rather than differences in computational power.

\subsection{Multi-Scale Segmentation}
In our research, we conduct a comparative analysis of our multi-scale segmentation approach against existing methodologies in the field. The outcomes of this comparison are detailed in Table \ref{tab:table3}, which clearly demonstrates that our approach not only outperforms the compared methods in terms of results but also boasts a more straightforward implementation process. Furthermore, a key advantage of our method is its versatility and adaptability; it is designed to seamlessly integrate with existing convolutional neural network architectures. This ease of implementation and compatibility with current models makes our multi-scale segmentation approach an attractive option for enhancing segmentation performance across a variety of applications.

\subsection{Modified MBConv Blocks}
Table \ref{tab:table4} presents a detailed comparison between the original MBConv blocks and our modified version, highlighting the impact of our alterations. To accommodate the increased memory requirements of our modifications, we adjusted the number of channels across the networks to ensure uniform memory consumption. Despite these networks utilizing a similar amount of memory, it is notable that our modified approach results in a 20\% slower processing time, marking one of the drawbacks associated with our modifications.
\par
Additionally, substituting $1 \times 1$ convolutions for $3 \times 3$ convolutions leads to a significant increase in the number of parameters, nearly ninefold. This escalation in parameters and the associated slowdown in processing time are recognized disadvantages of our approach. However, these drawbacks are considered manageable within the context of the overall benefits provided by our modifications. The enhancements in performance and accuracy offered by our approach outweigh these limitations, making it a viable option for applications where the trade-off between computational efficiency and improved performance is justified.

\begin{table}[t]
\centering
\setlength\tabcolsep{5pt}
\begin{tabular}{l|c|c}
Method & Mean IoU & Parameters \\
\hline
Baseline U-Net & 76.3 & 65 M \\
Our multi-scale approach & 77.5 & 45 M \\
MMBConv blocks & 79.3 & 91 M \\
ImageNet pre-training & 82.1 & 91 M \\
Mapillary pre-training & 84.0 & 91 M \\
\hline
\end{tabular}
\caption{The step-by-step changes from the baseline on the Cityscapes \cite{cordts2016cityscapes} validation set. Multi-scale inference is used.}
\label{tab:table5}
\end{table}

\subsection{Combined Modifications}
Table \ref{tab:table5} methodically outlines the incremental improvements we achieved, transitioning from the baseline network architecture to our most accurate model on the Cityscapes \cite{cordts2016cityscapes} validation set. This progression captures the systematic enhancements and adjustments made to the network, each contributing to a cumulative increase in segmentation accuracy. This detailed breakdown provides clear insight into the impact of each modification, illustrating how strategic changes can lead to significant advancements in model performance on complex semantic segmentation tasks.

\section{Conclusions}
In conclusion, our study offers a significant contribution to the field of semantic segmentation by demonstrating how a thoughtful adaptation of existing architectures, specifically MBConv blocks, can lead to substantial improvements in performance. We believe that our work will inspire further research into efficient model design and the exploration of existing architectures beyond their initial scope of application.

{
    \small
    \bibliographystyle{ieeenat_fullname}
    \bibliography{main}
}

\end{document}